%% file: iclr2025_conference.tex
\documentclass{article} 
\usepackage{iclr2025_conference,times}

\input{math_commands.tex}

\usepackage{hyperref}
\usepackage{url}
\usepackage{amssymb}
\usepackage{booktabs} 
\usepackage{booktabs}
\usepackage{multirow}
\usepackage[table]{xcolor}
\usepackage{graphicx}
\usepackage{arydshln}
\usepackage{tabularx}
\usepackage{float}
\usepackage{wrapfig}
\usepackage{caption}

\definecolor{grey}{HTML}{EEEEEE}
\definecolor{headergray}{HTML}{EFEFEF}
\definecolor{groupgray}{HTML}{F6F6F6}
\definecolor{basicblue}{HTML}{EEF4FA}
\definecolor{ragpink}{HTML}{FCEAEA}
\definecolor{embyellow}{HTML}{FFF4D8}
\definecolor{llmgreen}{HTML}{EDF7EA}
\definecolor{oursgreen}{HTML}{DDEFD7}

\definecolor{retrievalblue}{HTML}{F0F7FF}
\definecolor{oursblue}{HTML}{DDEEFF}
\definecolor{rerankpurple}{HTML}{F7F1FF}
\definecolor{ourspurple}{HTML}{E9DDFF}
\definecolor{oursbluetext}{HTML}{3B73B9}
\definecolor{ourspurpletext}{HTML}{7651B5}

\usepackage{amsmath}
\usepackage{algorithm}
\usepackage{algpseudocode}
\definecolor{sgblue}{RGB}{52,65,145}
\definecolor{sgorange}{RGB}{160,76,25}
\definecolor{sgpurple}{RGB}{112,45,145}
\definecolor{sggreen}{RGB}{20,120,65}

\usepackage[most]{tcolorbox}
\definecolor{retrievalblue}{HTML}{EFF7FE}
\definecolor{rerankpurple}{HTML}{F5F0FB}
\definecolor{retrievalpromptframe}{HTML}{426D98}
\definecolor{rerankpromptframe}{HTML}{735791}
\tcbset{appendixprompt/.style={enhanced,boxrule=0.7pt,arc=2pt,
  left=8pt,right=8pt,top=5pt,bottom=7pt,
  fontupper=\small,fonttitle=\bfseries\normalsize,
  coltitle=white,before skip=6pt,after skip=6pt}}

\newcommand{\best}[1]{\textbf{#1}}
\newcommand{\second}[1]{\underline{#1}}

\title{SkillGATE: Gate-Aware Monte Carlo Tree Search for Skill Retrieval}

\author{
  $^\dagger$Rongchen Zhao$^{1*,2}$ \quad
  $^\dagger$Yu Chen$^{2}$ \quad
  Yanming Yang$^{3}$ \quad
  Shijia Xu$^{1*}$ \\[0.3em]
  \bfseries
  Juyuan Wang$^{2}$ \quad
  Jin Xu$^{1}$ \quad
  Zibin Zheng$^{1}$ \quad
  $^{\ddagger}$ Jingping Liu$^{1}$ \\[0.5em]
  $^{1}$School of Software Engineering, Sun Yat-Sen University, Guangdong, China\\
  $^{2}$School of Future Technology, South China University of Technology, Guangdong, China \\
  $^{3}$Institute for Math and AI, Wuhan University, Wuhan, China \\
  \texttt{\{edwinzhaorc, cyu94987\}@gmail.com} \\
  \texttt{liujp68@mail.sysu.edu.cn}
}

\begin{document}

\maketitle
\begingroup
\renewcommand{\thefootnote}{\fnsymbol{footnote}}
\footnotetext[1]{Intern.\quad
 $^{\dagger}$Equal contribution.\quad
  $^{\ddagger}$Corresponding author.}
\endgroup
\begin{abstract}

Skill Retrieval (SR) aims to identify the most relevant skills from external skill libraries, and becomes increasingly challenging as libraries grow in scale and diversity. Existing methods either rank skills independently or rely on predefined graph propagation and hierarchical routing, making them vulnerable to semantic distractors, local trapping, and early routing errors. We formulate SR as an adaptive information-foraging process that coordinates region-level navigation with skill-level selection according to the utility and uncertainty observed during search. Based on this formulation, we propose SkillGATE, a graph-guided hierarchical retrieval framework with Gate-Aware Monte Carlo Tree Search (MCTS). SkillGATE constructs a graph-preserving hierarchical index and performs adaptive retrieval through selection, expansion, simulation, and backpropagation. G-PUCT guides action selection, expansion explores new regions, simulation evaluates candidate skills, and backpropagation updates search statistics. Experiments on six SR benchmarks show that SkillGATE consistently improves diverse retrieval and reranking backbones, achieving a 16.3\% improvement in overall R@1 over the strongest retriever-based baseline. Our code is available at \href{https://github.com/Edwinbe/SkillGATE-v1/}{here}.

\end{abstract}

\section{Introduction}
Skill Retrieval aims to identify the most relevant skill from an external skill library for a given query. As skill libraries grow to tens of thousands of entries, directly including all skills in the agent prompt becomes impractical and introduces irrelevant information. Effective SR is therefore essential for accurately retrieving the skill that best matches the query.

\begin{figure}[t]
\includegraphics[width=\columnwidth]{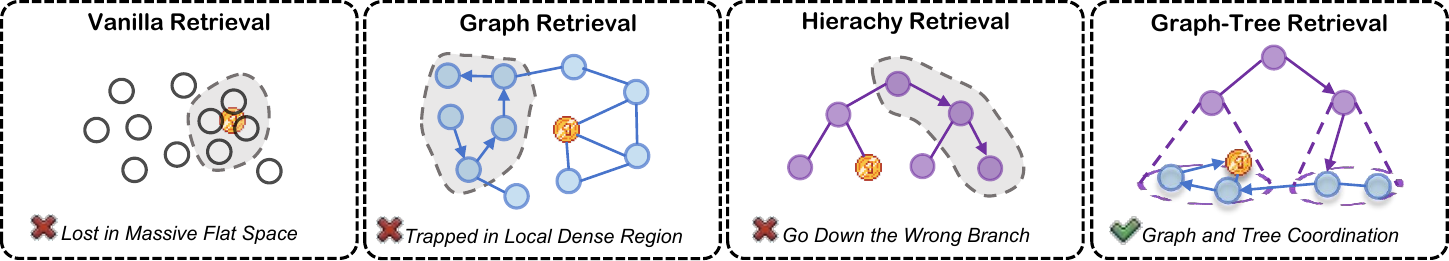}
\caption{Comparison between different skill retrieval paradigms.}
\label{fig:paradigm}
\vspace{-18pt}
\end{figure}

Existing approaches broadly fall into two paradigms: Vanilla SR and Structured SR. Vanilla SR treats skills as independent retrieval units and ranks them by query relevance using lexical matching or learned representations. Representative approaches include lexical retrievers~\citep{jones1972statistical,robertson2009bm25}, general-purpose dense retrievers~\citep{karpukhin2020dense,xiao2023cpack}, and embedding models adapted to skill retrieval through task-specific training~\citep{zheng2026skillrouter,cho2026skillret,wang2026skillnotdocument}. Structured SR organizes the library into an explicit index structure that guides retrieval through relations or groups of skills. Two common forms are \emph{graph indexing}~\citep{liu2026graphofskills,zeng2026goskills}, which represents inter-skill relations as edges and propagates relevance over the resulting graph, and \emph{hierarchical indexing}~\citep{zeng2026goskills, li2026organizing}, which organizes skills into groups and retrieves them through progressively finer routing decisions. Figure~\ref{fig:paradigm} provides an intuitive comparison of these retrieval paradigms and their characteristic search behaviors.

However, these approaches exhibit distinct limitations in navigating the skill space. Vanilla SR relies on independent relevance scores, making it susceptible to semantically similar distractors in a large flat search space. Graph-based SR exploits inter-skill relations, but propagation can become confined to local neighborhoods around the initial seeds. Hierarchy-based SR progressively narrows the search space, yet an early routing error may exclude relevant skills with limited opportunities for recovery. These limitations share a common cause: each paradigm operates within a largely fixed search regime and offers limited flexibility to revise its search scope based on evidence gathered during retrieval.
Effective SR therefore requires an adaptive search policy that coordinates exploitation of promising neighborhoods, exploration of alternative regions, and pruning of unproductive branches. This motivates viewing SR as adaptive information foraging, with the utility and uncertainty of observed returns guiding dynamic adjustments to search scope and strategy.

Motivated by Optimal Foraging Theory ~\citep{charnov1976optimal,pyke1977optimal}, we formulate skill retrieval as an adaptive search process over a structured skill space. In this view, different regions of the skill space are treated as \emph{patches}, with individual skills as resources within them. Retrieval uses the observed utility and uncertainty of the current region to guide whether the search should continue locally or move to alternative regions, while prioritizing promising skills within the explored region. This perspective directly addresses the rigidity of existing SR paradigms: rather than committing to a vanilla ranking, local propagation process, or hierarchical routing path, retrieval can dynamically adjust its search scope according to the utility and uncertainty of regions. This enables coordination between region-level navigation and skill-level selection as retrieval evidence accumulates.

Based on this formulation, we propose \textsc{SkillGATE}, a graph-guided hierarchical retrieval framework with gate-aware Monte Carlo Tree Search. During offline indexing, \textsc{SkillGATE} constructs a weighted skill graph by integrating semantic, lexical, and structured-label relations, and recursively partitions it into a hierarchical index while preserving skill-level graph connections. This structure supports both progressive region-level navigation and cross-region transitions. During online retrieval, we develop a gate-aware MCTS algorithm that coordinates region-level navigation with skill-level selection through node-aware descent, skill-aware descent, horizontal search, and pruning. Its G-PUCT policy jointly considers historical return, normalized return entropy, query-relevance priors, and visit statistics to adapt the search strategy as retrieval evidence accumulates. Expansion explores promising unvisited regions, simulation evaluates their contained skills and collects relevant candidates, and backpropagation separately updates search visitation and hierarchical reward statistics. Overall, these components enable \textsc{SkillGATE} to adaptively navigate the structured skill space while maintaining fine-grained relevance to the query.

Our primary contributions are summarized as follows. 1) We formulate skill retrieval as an adaptive information-foraging process that coordinates region-level navigation with skill-level selection. 2) We propose \textsc{SkillGATE}, a graph-guided hierarchical retrieval framework with gate-aware MCTS for adaptive search and cross-region exploration. 3) Extensive experiments on six SR benchmarks show that \textsc{SkillGATE} consistently improves diverse retrieval and reranking backbones, achieving a 16.3\% improvement in overall R@1 over the strongest retriever-based baseline.

\section{Related Work} 

\textbf{Skill Retrieval.}
Existing SR methods can be broadly divided into two paradigms: Vanilla SR, which ranks skills independently through query–skill matching, and Structured SR, which exploits explicit organization or relations among skills.
Vanilla SR mainly follows retrieve-and-rerank pipelines. SkillFlow~\citep{li2026skillflow} combines dense retrieval, cross-encoder reranking, and LLM-based selection. SkillRouter~\citep{zheng2026skillrouter} leverages full skill content with false-negative filtering and listwise reranking. SkillRet~\citep{cho2026skillret} provides large-scale benchmarks and retrieval-oriented training data, while R3~\citep{wang2026skillnotdocument} uses rejection signals to supervise query-conditioned skill compatibility. Such methods largely operate over independently scored skills or retrieved shortlists.
Structured SR explicitly organizes the skill library. AgentSkillOS~\citep{li2026organizing} builds a capability tree for coarse-to-fine retrieval; Graph-of-Skills~\citep{liu2026graphofskills} performs dependency-aware retrieval over a skill graph; and GoSkills~\citep{zeng2026goskills} organizes skills into graph-connected groups for structured retrieval and context assembly. While these methods demonstrate the value of structural organization, they typically rely on predefined routing or propagation mechanisms. In contrast, SkillGATE combines hierarchical organization with graph connectivity and adaptively coordinates region-level navigation with skill-level selection.

\textbf{Document Retrieval.}
Document retrieval approaches can be broadly categorized into flat retrieval, which scores individual documents or passages, and structured retrieval, which exploits hierarchies or relational graphs. Early work on flat retrieval primarily relied on lexical methods, such as BM25~\citep{robertson2009bm25} and TF-IDF~\citep{jones1972statistical}, which use vocabulary-based term statistics to estimate relevance through weighted term overlap between queries and documents. More recent efforts have increasingly explored dense retrievers, including Qwen3-Embedding~\citep{zhang2025qwen3embedding} and BGE~\citep{chen2024bge}. These methods encode queries and documents into a shared dense vector space and rank candidates by embedding similarity, enabling semantic matching beyond exact lexical overlap.
Recent efforts along structured retrieval mainly includes: HippoRAGv2~\citep{gutierrez2025ragmemory}, which integrates passages and LLM guidance into Personalized PageRank (PPR)~\citep{jeh2003scaling} retrieval; ComoRAG~\citep{wang2026comorag}, which constructs a hierarchical knowledge index over raw passages, semantic abstractions, and narrative summaries to support iterative multi-layer retrieval; PonsRAG~\citep{zhao2026ponsrag}, which connects narrative knowledge across layers through graph-based indexing and coordinates cross-layer retrieval to recover complementary evidence.

\label{headings}
\begin{figure}
\includegraphics[width=\columnwidth]{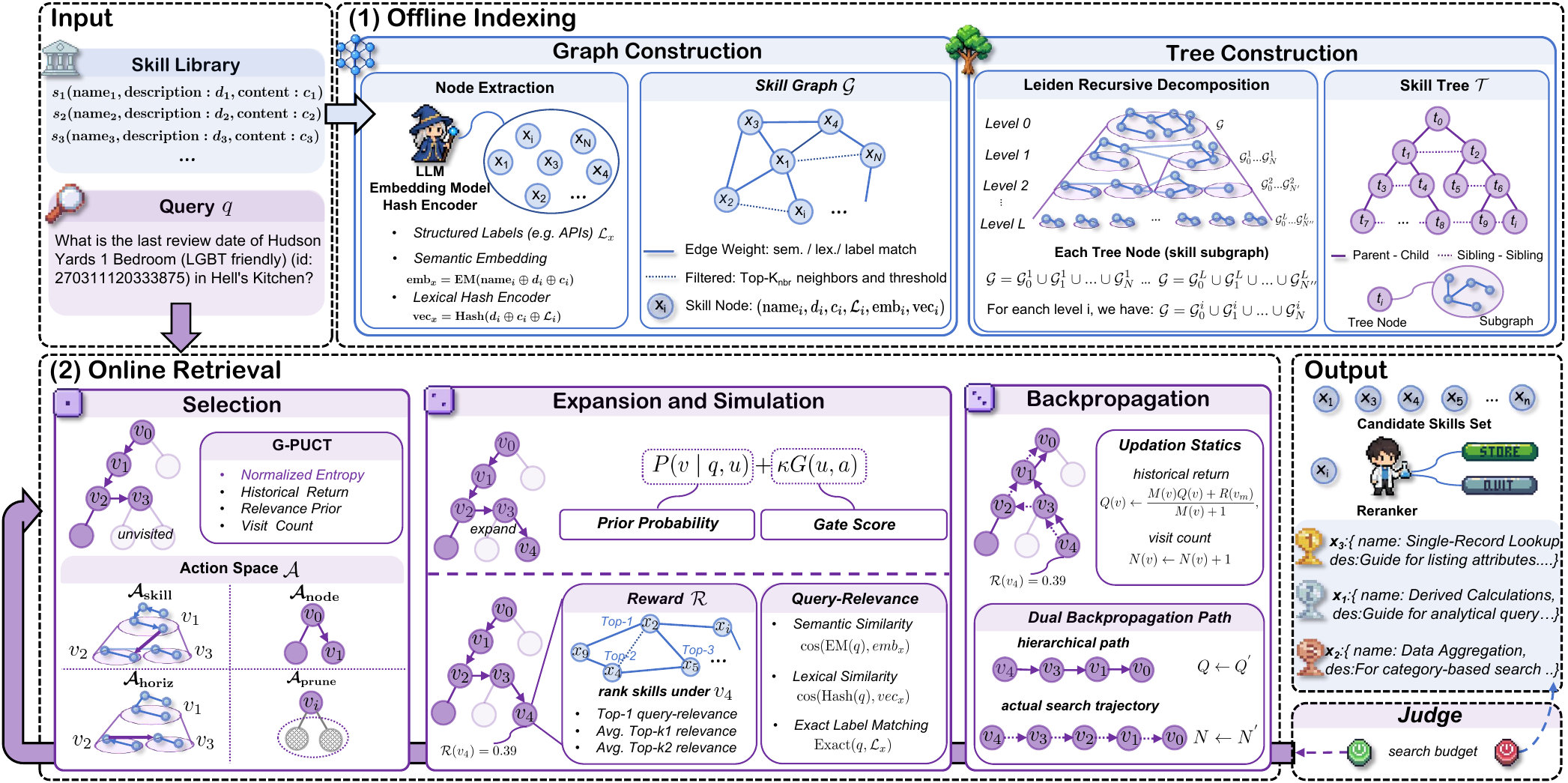}
\caption{Framework Overview. }
\label{fig:framework}
\vspace{-18pt}
\end{figure}

\section{Overview}
In this section, we first formalize the skill retrieval task and then present an overview of our proposed framework, \textsc{SkillGATE}.

\textbf{Problem formulation.}
Let $\mathcal{S}=\{s_1,s_2,\ldots,s_N\}$ denote a library of $N$ skills. Each skill $s_i$ is represented by a tuple $(\mathrm{name}_i,d_i,c_i)$, comprising its name, functional description, and detailed execution content, respectively. Given a query $q$, skill retrieval aims to identify the top-$K$ skills that are most relevant to the query and best suited to its functional requirements.
Let $F$ denote an SR framework that takes $q$ and $\mathcal{S}$ as input and returns a ranked list of skills. The resulting top-$K$ skill set is given by
\begin{equation}
\mathcal{S}^{*}_K(q)
=
\operatorname{TopK}\!\left(F(q,\mathcal{S})\right).
\end{equation}

\textbf{Our framework.}
As illustrated in Figure~\ref{fig:framework}, \textsc{SkillGATE} consists of \textit{offline indexing} and \textit{online retrieval}.
During offline indexing, we first construct a weighted skill graph $\mathcal{G}$, whose nodes represent skills and whose edge weights capture inter-skill relevance. We then recursively partition $\mathcal{G}$ to build a hierarchy in which each non-root node corresponds to a subgraph of $\mathcal{G}$. The resulting index organizes the skill library hierarchically while preserving graph connections to support retrieval both within and across subgraphs.
Given a query $q$, online retrieval employs our gate-aware MCTS to explore the hierarchy and retrieve relevant skills through four stages: selection, expansion, simulation, and backpropagation.
Starting from the root, \textit{selection} repeatedly applies the proposed G-PUCT policy until reaching an expandable node, balancing high-reward and low-entropy directions with exploration of less-visited alternatives.
\textit{Expansion} adds an unvisited child or sibling node. From the expanded node, \textit{simulation} follows the default policy until termination or budget exhaustion. Let $v_m$ denote the final node, where the relevance function $r$ evaluates contained skills to produce reward $R(v_m)$, and highly relevant skills are added to the candidate set $\mathcal{S}^{*}(q)$.
During \textit{backpropagation}, we decouple visit-count updates from reward assignment. Visit counts are updated along the search trajectory, while $R(v_m)$ is propagated through the hierarchical path to update node rewards $Q$.
After the prescribed number of iterations, \textsc{SkillGATE} returns the top-$K$ skills from $\mathcal{S}^{*}(q)$.

\section{Methodology}
\subsection{Offline Indexing}
\textbf{Graph construction.}
Given a skill $s_i \in \mathcal{S}$, we use an LLM to extract a structured label set $\mathcal{L}_i$ from its tuple representation $(\mathrm{name}_i,d_i,c_i)$. These labels capture the skill's input and output fields, as well as the tools (e.g., APIs) involved in its execution. We then compute a semantic embedding and a lexical hash vector: $\mathrm{emb}_i = \operatorname{EM}(\mathrm{name}_i \oplus d_i \oplus c_i)$ and $\mathrm{vec}_i = \operatorname{Hash}(d_i \oplus c_i \oplus \mathcal{L}_i)$, where $\oplus$ denotes concatenation. Each skill is represented as a graph node
$x_i=(\mathrm{name}_i,d_i,c_i,\mathcal{L}_i,\mathrm{emb}_i,\mathrm{vec}_i)\in\mathcal{X}$.

For each pair of skill nodes, we compute an edge weight $w_{ij}$ as a weighted combination of semantic similarity $\cos(\mathrm{emb}_i,\mathrm{emb}_j)$, lexical similarity $\cos(\mathrm{vec}_i,\mathrm{vec}_j)$, and label consistency $\operatorname{Jaccard}(\mathcal{L}_i,\mathcal{L}_j)$. To maintain graph sparsity, we retain, for each node, at most $K_{\mathrm{nbr}}$ neighbors with the highest edge weights among those exceeding a threshold $\tau_e$. The resulting weighted skill graph is denoted by $\mathcal{G}=(\mathcal{X},\mathcal{E})$.

\textbf{Tree construction.}
We construct a hierarchical index tree $\mathcal{T}$ whose nodes are subgraphs of the skill graph $\mathcal{G}$, with $\mathcal{G}$ itself as the root. Starting from the root, we recursively apply the Leiden community detection algorithm~\citep{traag2019louvain} to partition each subgraph into communities, each forming a child node in $\mathcal{T}$. Recursion stops when a node reaches the maximum decomposition depth $l_{\max}$ or contains fewer than $n_{\min}$ skills.

Each tree node $t$ retains the original skill-level edges between its member skills. To obtain node-level representations, we average the semantic embeddings and lexical hash vectors of these skills and normalize the resulting mean vectors, yielding $\mathrm{emb}_t$ and $\mathrm{vec}_t$, respectively. We also merge their label sets into a structured label set $\mathcal{L}_t$. Each node is thus a subgraph augmented with the aggregated features $(\mathrm{emb}_t,\mathrm{vec}_t,\mathcal{L}_t)$.

\subsection{Online Retrieval}
We now describe the online retrieval procedure of \textsc{SkillGATE}, which consists of selection, expansion, simulation, and backpropagation.

\subsubsection{Selection}
Starting from the root, the selection stage traverses the current MCTS tree by repeatedly choosing the most promising action until an expandable node is reached or a pruning action is selected. At each node $u$, the search considers four types of actions.
\textit{Node-aware descent} ($\mathcal{A}_{\mathrm{node}}$) moves to a child node to further narrow the retrieval scope.
\textit{Skill-aware descent} ($\mathcal{A}_{\mathrm{skill}}$) performs PPR within the current subgraph $u$ to identify the highest-scoring skill, and then moves to the child node containing that skill.
\textit{Horizontal search} ($\mathcal{A}_{\mathrm{horiz}}$) moves to a sibling node in the index hierarchy through cross-subgraph skill edges, allowing the search to leave a potentially suboptimal branch.
\textit{Pruning} ($\mathcal{A}_{\mathrm{prune}}$) excludes the current node and its descendants from further exploration during the current iteration and immediately initiates simulation for reward evaluation.
The global action space is therefore defined as
$\mathcal{A}
=
\mathcal{A}_{\mathrm{node}}
\cup
\mathcal{A}_{\mathrm{skill}}
\cup
\mathcal{A}_{\mathrm{horiz}}
\cup
\mathcal{A}_{\mathrm{prune}}$.
More details are provided in Appendix \ref{app:act}.

To guide action selection, we introduce G-PUCT, which incorporates four signals: historical return, normalized entropy, a query-relevance prior, and visit counts.
The historical return $Q$ (Section~\ref{sec:back})reflects the past retrieval performance of a search direction and is normalized to $[0,1]$.
The normalized entropy $H$ characterizes the concentration of a visit-weighted distribution over visited child nodes.
The prior probability $P$ (Section~\ref{sec:expand}) measures the relevance of a candidate node to the query, while the visit counts encourage exploration of less-visited candidates.
For an action $a$ that leads from the current node $u$ to a candidate node $v$, the G-PUCT score is defined as
\begin{equation}
\operatorname{G-PUCT}(a,v\mid u)
={}\bigl(1-\kappa(u)\bigr)Q(v)
+
\kappa(u)G(u,a) +
\lambda P(v)\frac{\sqrt{\max(1,N(u))}}{1+N(v)},
\end{equation}
where $N(\cdot)$ denotes the visit count, $G(u,a)$ is an action-specific gate score, and $\lambda$ controls the strength of prior-guided exploration.
The mixing coefficient $\kappa(u)$ is defined as
\begin{equation}
\kappa(u)
=
1-\exp\!\left(-\frac{N(u)}{\tau_{\kappa}}\right),
\end{equation}
where $\tau_{\kappa}>0$ controls how quickly the weighting shifts from the candidate node's historical return $Q(v)$ toward the gate score $G(u,a)$.
A larger $\tau_{\kappa}$ slows this transition, preserving the direct contribution of $Q(v)$ over more visits.
For pruning actions, we assign sufficiently small values to the return and prior components of the score.

The gate score combines the normalized historical return $Q(u)$ and entropy $H(u)$ as follows:
\begin{equation}
G(u,a)=
\begin{cases}
Q(u)H(u),
& a \in \mathcal{A}_{\mathrm{node}}, \\[4pt]
Q(u)\bigl(1-H(u)\bigr),
& a \in \mathcal{A}_{\mathrm{skill}}, \\[4pt]
\bigl(1-Q(u)\bigr)H(u),
& a \in \mathcal{A}_{\mathrm{horiz}}, \\[4pt]
\bigl(1-Q(u)\bigr)\bigl(1-H(u)\bigr),
& a \in \mathcal{A}_{\mathrm{prune}}.
\end{cases}
\end{equation}
To compute $H(u)$, let $\mathcal{C}(u)$ denote the set of visited children of $u$.
When $\mathcal{C}(u)$ is nonempty, we define a probability distribution over these children by weighting their exponentiated mean returns by their visit counts:
\begin{equation}
p(v_i\mid u)
=
\frac{
N(v_i)\exp\!\left(Q(v_i)/\tau_H\right)
}{
\sum_{v_j\in\mathcal{C}(u)}
N(v_j)\exp\!\left(Q(v_j)/\tau_H\right)
},
\end{equation}
where $\tau_H>0$ controls the influence of return differences on the distribution.
For nodes with at least two visited children, we normalize the Shannon entropy \citep{shannon1948mathematical} by its maximum value:
\begin{equation}
H(u)
=
-\frac{1}{\log |\mathcal{C}(u)|}
\sum_{v_i\in\mathcal{C}(u)}
p(v_i\mid u)\log p(v_i\mid u).
\end{equation}
A higher $H(u)$ indicates a more evenly distributed probability mass, whereas a lower value indicates greater concentration on a smaller subset of children.
For a single visited child, we set $H(u)=0$; when no child has been visited, we initialize $H(u)=1$ as a convention for the absence of observations.
These definitions ensure that $H(u)\in[0,1]$ and, together with $Q(u)\in[0,1]$, keep all four gate scores nonnegative.
For each feasible action, we compute the corresponding gate and G-PUCT scores and select the action with the highest G-PUCT score.

\subsubsection{Expansion and Simulation}
\label{sec:expand}

The expansion stage begins at an expandable node $u$ and adds an unvisited successor $v$ to the MCTS tree. The choice of $v$ is guided by its query-relevance prior $P(v \mid q, u)$ (also denoted as $P(v)$) and the gate score $\kappa(u)G(u,a)$ of the action $a$ leading to it.
To compute the prior $P(v \mid q, u)$, we first define the relevance $r(q,v)$ between the query $q$ and a candidate node $v$ as a weighted combination of semantic similarity $\cos(\operatorname{EM}(q),emb_v)$, lexical similarity $\cos(\operatorname{Hash}(q),vec_v)$, and exact label matching $\operatorname{Exact}(q,\mathcal{L}_v)$. Let $\mathcal{D}(u)$ denote the set of candidate successor nodes reachable from $u$ through feasible non-pruning actions, including eligible child and sibling nodes. We obtain the prior probability by applying a softmax over their relevance scores:
\begin{equation}
P(v\mid q, u)
=
\frac{
\exp\!\left(r(q,v)/\tau_P\right)
}{
\sum_{v'\in\mathcal{D}(u)}
\exp\!\left(r(q,v')/\tau_P\right)
},
\end{equation}
where $\tau_P>0$ controls the sharpness of the prior distribution. This distribution is defined over all candidate successors, while expansion is restricted to unvisited ones.

After expansion, the simulation samples feasible non-pruning transitions according to the prior $P(v\mid q,u)$ until no feasible transition remains or the search budget is reached.
The utility of this rollout is evaluated based on the relevance between the query $q$ and the skills contained in the subgraph node $v_m$. Specifically, we rank these skills using $r(q,x)$ and define the simulation reward $\mathcal{R}(v_m)$ as a weighted sum of three terms: the top-1 relevance score, the average relevance score over the top-$k_1$ skills, and the average relevance score over the top-$k_2$ skills. The top-$k_2$ skills are further added to a global candidate pool.

\subsubsection{Backpropagation}
\label{sec:back}
The backpropagation stage updates the statistics used in subsequent MCTS iterations while decoupling search visitation from reward assignment.
Let $\mathcal{V}_{\mathrm{visit}}$ denote the actual search trajectory of the current iteration, and let $\mathcal{V}_{\mathrm{hier}}$ denote the hierarchical path from the final node $v_m$ to the root.
The visit count is updated along the actual search trajectory:
$N(v)\leftarrow N(v)+1$, where $v\in\mathcal{V}_{\mathrm{visit}}$.
In contrast, the simulation reward $R(v_m)$ is propagated along $\mathcal{V}_{\mathrm{hier}}$ to update the historical mean return $Q$. Since reward assignment is independent of node visitation, we maintain a separate reward count $M(v)$. For each $v\in\mathcal{V}_{\mathrm{hier}}$, the update is
\begin{equation}
\begin{aligned}
Q(v)
&\leftarrow
\frac{M(v)Q(v)+R(v_m)}{M(v)+1}, \ \ \ \ M(v) \leftarrow M(v)+1.
\end{aligned}
\end{equation}
Visit and reward counts are initialized as $N(v)=M(v)=0$. For nodes without observed rewards, we use the neutral calibrated value $Q(v)=0.5$.
This dual-path update mechanism allows $N$ to record the actual exploration process, while $Q$ attributes the retrieved reward to the hierarchical branch producing the relevant skills and its ancestors.

\section{Experiment}
In this section, we conduct extensive experiments to evaluate the effectiveness of our method, followed by several analyses to further investigate its key components and behaviors.

\subsection{Experimental Setup}


\textbf{Benchmarks.}
We evaluate all methods on six benchmarks: TheoremQA~\citep{chen2023theoremqa}, LogicBench~\citep{parmar2024logicbench}, ToolQA~\citep{zhuang2023toolqa}, CHAMP~\citep{mao2024champ}, MedCalc-Bench~\citep{khandekar2024medcalc}, and BigCodeBench~\citep{zhuo2025bigcodebench}. All queries are evaluated against the shared 26,262-skill library from SRA-Bench~\citep{su2026skillretrieval}. Following SRA-Bench, we report Recall@1 (R@1) and Recall@10 (R@10) for each benchmark, along with their averages across all benchmarks.

\textbf{Baselines.}
We compare against two categories of baselines: retriever-based and reranker-based methods. Retriever-based baselines include general retrievers (BGE-Emb~\citep{chen2024bge}, Ada-002, and BM25~\citep{robertson2009bm25}), structured retrievers (GoS~\citep{liu2026graphofskills}, GoSkills~\citep{zeng2026goskills}, SkillFlow~\citep{li2026skillflow}, and AgentSkillOS~\citep{li2026organizing}), and fine-tuned retrievers (SkillRouter-Emb~\citep{zheng2026skillrouter}, SkillRet-Emb~\citep{cho2026skillret}, and R3-Emb~\citep{wang2026skillnotdocument}). Reranker-based baselines augment retrievers with reranking models, including general rerankers (GPT-4o-mini, Qwen3-4B~\citep{yang2025qwen3}, and BGE-Rerank~\citep{chen2024bge}), structured retrieval methods with reranking, and fine-tuned rerankers from SkillRouter~\citep{zheng2026skillrouter} and R3~\citep{wang2026skillnotdocument}. For structured methods without an original reranker, we use GPT-4o-mini; otherwise, we follow their original settings. Implementation details are provided in Appendix~\ref{app:details}.

\begin{table*}[t]
\centering
\scriptsize
\setlength{\tabcolsep}{2.0pt}
\renewcommand{\arraystretch}{1.06}
\caption{Main results (\%) on six benchmarks for SR. \textbf{Blue} denotes retriever-based methods and \textbf{purple} denotes reranker-based methods. \textbf{Bold} marks the best in ours, \underline{underline} marks the best baseline, and \textbf{\textit{Improvement}} marks their performance gap.}
\label{tab:main_results}
\begin{tabularx}{\textwidth}{l*{14}{>{\centering\arraybackslash}X}}
\toprule
\multicolumn{1}{c}{\multirow{2}{*}{\textbf{Method}}}
& \multicolumn{2}{c}{\textbf{TheoremQA}}
& \multicolumn{2}{c}{\textbf{LogicBench}}
& \multicolumn{2}{c}{\textbf{ToolQA}}
& \multicolumn{2}{c}{\textbf{CHAMP}}
& \multicolumn{2}{c}{\textbf{MedCalc.}}
& \multicolumn{2}{c}{\textbf{BigCode.}}
& \multicolumn{2}{c}{\textbf{Overall}} \\
\cmidrule(lr){2-3}
\cmidrule(lr){4-5}
\cmidrule(lr){6-7}
\cmidrule(lr){8-9}
\cmidrule(lr){10-11}
\cmidrule(lr){12-13}
\cmidrule(lr){14-15}
& \textbf{R@1} & \textbf{R@10}
& \textbf{R@1} & \textbf{R@10}
& \textbf{R@1} & \textbf{R@10}
& \textbf{R@1} & \textbf{R@10}
& \textbf{R@1} & \textbf{R@10}
& \textbf{R@1} & \textbf{R@10}
& \textbf{R@1} & \textbf{R@10} \\
\midrule

\rowcolor{retrievalblue}
\multicolumn{15}{c}{\textit{\textbf{Retriever-based Methods}}} \\
\rowcolor{retrievalblue}
BM25       & 57.2 & 80.7 & 12.0 & 36.1 &  7.0 & 55.1 & 13.2 & 36.1 & 29.3 & 69.2 & \second{23.6} & 61.1 & 23.0 & 59.3 \\
\rowcolor{retrievalblue}
BGE-Emb & 66.8 & 86.1 &  4.1 & 20.5 & 32.2 & 83.4 &  9.8 & 34.0 & 41.4 & 70.1 & 20.7 & 62.1 & 31.6 & 65.7 \\
\rowcolor{retrievalblue}
Ada-002 & \second{68.0} & \second{90.8} & 9.1 & 31.6 & 18.8 & 54.8 & 13.7 & 41.7 & 82.6 & 98.4 & 18.2 & 52.0 & 36.9 & 64.2 \\

\addlinespace[1pt]
\hdashline
\addlinespace[1pt]
\rowcolor{retrievalblue}
GoS        & 54.8 & 83.4 &  7.6 & 25.5 &  6.8 & 20.4 &  4.9 & 25.1 & 67.1 & 81.8 & 17.2 & 49.3 & 28.0 & 48.6 \\
\rowcolor{retrievalblue}
GoSkills   & 36.0 & 51.1 &  6.2 & 10.9 &  7.0 & 30.8 &  5.0 &  8.6 & 13.7 & 25.4 & 16.3 & 36.6 & 14.1 & 30.0 \\
\rowcolor{retrievalblue}
SkillFlow (emb)  & 58.8 & 82.1 &  1.1 &  8.0 & 15.4 & 51.3 &  7.7 & 26.6 & 32.6 & 60.5 & 13.4 & 41.7 & 22.2 & 48.3 \\

\addlinespace[1pt]
\hdashline
\addlinespace[1pt]
\rowcolor{retrievalblue}
SkillRouter-Emb-0.6B & 56.1 & 85.4 &  5.4 & 30.7 & 10.0 & 34.6 & \second{17.3} & \second{51.3} & 71.4 & 94.9 & 11.7 & 39.4 & 28.9 & 55.1 \\
\rowcolor{retrievalblue}
SkillRet-Emb-0.6B    & 61.3 & 82.9 & 13.8 & 56.8 & 26.9 & 69.1 &  6.9 & 24.6 & \second{96.5} & \second{100.0} & 22.6 & 68.9 & \second{42.3} & 73.7 \\
\rowcolor{retrievalblue}
R3-Emb-0.6B   & 32.5 & 75.0 & \second{30.3} & \second{84.2} & \second{34.5} & \second{97.9} & 1.1 & 20.0 & 92.3 & 99.8 & 22.6 & \second{71.2} & 41.5 & \second{84.3} \\

\midrule
\rowcolor{oursblue}
\multicolumn{15}{l}{\textbf{Ours}} \\
\rowcolor{oursblue}
\textit{+Ada-002}
& 74.7 & 93.8 & 13.8 & 54.0 & 41.4 & 68.2 & 17.8 & 56.1 & 93.6 & \best{98.1} & 23.0 & 61.1 & 47.9 & 73.8 \\
\rowcolor{oursblue}
\textit{+SkillRet-Emb-0.6B}
& \best{78.6} & 96.4 & 20.5 & 74.2 & 62.5 & 90.4 & \best{22.7} & 61.6 & 93.6 & 95.3 & 26.1 & 64.6 & 55.8 & 83.3 \\

\rowcolor{oursblue}
\textit{+R3-Emb-0.6B}
& 77.9 & \best{97.2} & \best{33.2} & \best{87.1} & \best{65.9} & \best{92.9} & 22.2 & \best{62.6} & \best{94.1} & 97.5 & \best{26.5} & \best{72.8} & \best{58.6} & \best{88.1} \\
\midrule
\rowcolor{oursblue}
\textit{\textbf{Improvement}}
& \textit{+10.6} & \textit{+6.4}
& \textit{+2.9} & \textit{+2.9}
& \textit{+31.4} & \textit{-5.0}
& \textit{+5.4} & \textit{+11.3}
& \textit{-2.4} & \textit{-1.9}
& \textit{+2.9} & \textit{+1.6}
& \textit{+16.3} & \textit{+3.8} \\
\midrule
\rowcolor{rerankpurple}
\multicolumn{15}{c}{\textit{\textbf{Reranker-based Methods}}} \\
\rowcolor{rerankpurple}
Ada-002+GPT-4o-mini & 81.3 & \second{93.6} & 20.4 & 35.9 & 33.7 & 60.0 & 24.9 & 52.7 & \second{97.2} & 99.1 & 19.0 & 57.3 & 47.9 & 68.3 \\
\rowcolor{rerankpurple}
Ada-002+Qwen3-4b & 77.1 & 92.8 & 17.8 & 35.8 & 29.8 & 59.8 & 19.3 & 51.6 & 87.5 & 98.5 & 19.9 & 56.5 & 43.9 & 67.8 \\
\rowcolor{rerankpurple}
BGE-Emb+BGE-Reranker & 66.4 & 86.9 & 10.8 & 18.2 & 39.2 & 86.9 & 7.0 & 28.0 & 22.3 & 53.7 & 7.2 & 36.7 & 27.4 & 57.4 \\

\addlinespace[1pt]
\hdashline
\addlinespace[1pt]
\rowcolor{rerankpurple}
GoS+GPT-4o-mini        & 75.4 & 83.4 & 16.2 & 27.9 & 14.7 & 23.2 & 16.4 & 27.4 & 81.2 & 82.4 & \second{23.8} & 50.8 & 38.8 & 50.2 \\
\rowcolor{rerankpurple}
GoSkills+GPT-4o-mini   & 50.5 & 54.2 &  7.4 & 10.9 & 25.1 & 30.7 &  7.0 &  8.6 & 25.4 & 25.4 & 23.3 & 36.6 & 25.1 & 30.4 \\
\rowcolor{rerankpurple}
SkillFlow  & 69.3 & 80.7 &  2.5 &  3.8 & \second{47.9} & 67.3 & 24.9 & 44.6 & 24.4 & 42.4 &  5.4 & 23.1 & 29.8 & 44.9 \\
\rowcolor{rerankpurple}
AgentSkillOS & 15.8 & 18.5 & 1.5 & 2.8 & 15.0 & 22.2 & 6.7 & 17.3 & 84.0 & 88.6 & 2.6 & 9.9 & 24.3 & 29.7 \\

\addlinespace[1pt]
\hdashline
\addlinespace[1pt]
\rowcolor{rerankpurple}
SkillRouter-Rerank-0.6B & \second{83.1} & 91.8 & 24.5 & 40.1 & 36.9 & 49.0 & \second{28.7} & \second{61.5} & 95.1 & 96.8 & 13.4 & 35.6 & 48.1 & 61.1 \\
\rowcolor{rerankpurple}
R3-Rerank-0.6B    & 78.7 & 85.1 & \second{39.1} & \second{90.7} & 34.5 & \second{97.9} & 20.3 & 37.5 & 91.2 & \second{99.7} & 18.7 & \second{59.5} & \second{48.9} & \second{84.9} \\

\midrule
\rowcolor{ourspurple}
\multicolumn{15}{l}{\textbf{Ours}} \\
\rowcolor{ourspurple}
\textit{+GPT-4o-mini}
& 82.9 & 95.2 & 23.0 & 60.0 & 48.5 & 68.0 & 27.5 & 64.0 & \best{97.7} & \best{99.1} & \best{24.9} & \best{66.3} & 53.9 & 76.5 \\
\rowcolor{ourspurple}
\textit{+SkillRouter-Rerank-0.6B}
& 86.1 & 96.1 & 29.6 & 78.2 & 62.8 & 90.1 & 30.6 & 69.4 & 93.5 & 95.3 & 15.3 & 51.1 & 56.2 & 81.2 \\

\rowcolor{ourspurple}
\textit{+R3-Rerank-0.6B}
& \best{87.7} & \best{97.3} & \best{39.9} & \best{93.3} & \best{72.5} & \best{92.4} & \best{37.0} & \best{70.5} & 85.0 & 97.6 & 18.7 & 59.9 & \best{59.7} & \best{86.5} \\
\midrule
\rowcolor{ourspurple}
\textit{\textbf{Improvement}}
& \textit{+4.6} & \textit{+3.7}
& \textit{+0.8} & \textit{+2.6}
& \textit{+24.6} & \textit{-5.5}
& \textit{+8.3} & \textit{+9.0}
& \textit{+0.5} & \textit{-0.6}
& \textit{+1.1} & \textit{+6.8}
& \textit{+10.8} & \textit{+1.6} \\
\bottomrule
\end{tabularx}
\end{table*}

\subsection{Main Results}

To evaluate the effectiveness of our framework, we compare it with two categories of baselines. From results shown in Table \ref{tab:main_results}, we make the following observations:
1) Our framework consistently outperforms existing methods. Under the retriever setting, our method achieves 58.6\% R@1 and 88.1\% R@10, improving over the strongest baselines by 16.3 and 3.8 percentage points, respectively. Under the reranker setting, it achieves 59.7\% R@1 and 86.5\% R@10, with gains of 10.8 and 1.6 points. Except for ToolQA and MedCalc., our method improves or matches the best baseline across all benchmarks and metrics.
2) The gains in R@1 are generally larger than those in R@10. The overall R@1 improves by 16.3 and 10.8 points under the retriever and reranker settings, respectively, while the corresponding R@10 gains are 3.8 and 1.6 points. This suggests that our framework is particularly effective at ranking the most relevant skill at the top. In particular, on ToolQA, the retriever and reranker settings improve R@1 by 31.4 and 24.6 points, respectively, while reducing R@10 by 5.0 and 5.5 points. This is because the candidate skills in ToolQA are highly similar, and our more focused branch selection improves top-1 precision but may miss the correct branch, thereby reducing Top-10 coverage.
3) Our framework generalizes well across different backbones. With Ada-002, SkillRet-Embedding, and R3-Embedding, our method achieves overall R@1/R@10 scores of 47.9/73.8, 55.8/83.3, and 58.6/88.1, respectively, all outperforming their corresponding embedding-based baselines. Similar gains are observed with GPT-4o-mini, SkillRouter-Reranker, and R3-Reranker, showing that the framework does not depend on a specific backbone.
4) ToolQA shows the largest R@1 gains, while CHAMP shows the largest R@10 gains. On ToolQA, R@1 improves by 31.4 and 24.6 points under the retriever and reranker settings, respectively, which is consistent with the need to distinguish among functionally similar tool skills. In contrast, CHAMP achieves the largest R@10 gains, with improvements of 11.3 and 9.0 points, suggesting better coverage of multiple relevant mathematical skills.


\subsection{Ablation Study}

To evaluate the contribution of each component in our framework, we conduct ablation studies using Ada-002 without rerankers. For the indexing ablations, the graph-only variant performs PPR directly over the skill graph, while the tree-only variant performs vanilla MCTS over the hierarchical tree. For the retrieval ablations, we remove each proposed component individually while keeping the remaining settings unchanged. Specifically, w/o Reward replaces the proposed reward with the average relevance over all skills in the final node; w/o $\mathcal{V}_{\mathrm{hier}}$ and w/o $\mathcal{V}_{\mathrm{visit}}$ collapse the decoupled backpropagation into a single update path; the remaining variants remove $\mathcal{A}_{\mathrm{skill}}$, $\mathcal{A}_{\mathrm{horiz}}$, or $\mathcal{A}_{\mathrm{prune}}$, or replace G-PUCT with vanilla PUCT. The results are reported in Table \ref{tab:ablation}.
Overall, all components contribute positively to performance. The full model achieves 47.9\% R@1 and 73.8\% R@10, while using only the graph or tree reduces performance to 29.9/45.3\% and 13.8/17.3\%, respectively, showing their complementary roles in narrowing the search space and capturing relevant skills beyond a single hierarchical path. Among the retrieval components, removing $\mathcal{A}_{\mathrm{skill}}$ causes the largest drop of 15.7/28.4 points, followed by $\mathcal{A}_{\mathrm{horiz}}$ with 6.6/8.8 points, highlighting the importance of skill-aware descent and horizontal transitions. Replacing G-PUCT with vanilla PUCT decreases R@1/R@10 by 5.3/7.4 points, confirming the benefit of the gate-aware selection strategy. The reward design and decoupled backpropagation also provide consistent gains, with removing the reward, $\mathcal{V}{\mathrm{hier}}$, and $\mathcal{V}{\mathrm{visit}}$ causing drops of 1.4/2.9, 1.9/4.9, and 2.2/5.5 points, respectively.


\begin{table*}[t]
\centering
\scriptsize
\setlength{\tabcolsep}{2.5pt}
\renewcommand{\arraystretch}{1.06}
\caption{
Ablation study on the indexing and retrieval components of our framework.
}
\label{tab:ablation}
\begin{tabularx}{\textwidth}{l*{14}{>{\centering\arraybackslash}X}}
\toprule
\multicolumn{1}{c}{\multirow{2}{*}{\textbf{Method}}}
& \multicolumn{2}{c}{\textbf{TheoremQA}}
& \multicolumn{2}{c}{\textbf{LogicBench}}
& \multicolumn{2}{c}{\textbf{ToolQA}}
& \multicolumn{2}{c}{\textbf{CHAMP}}
& \multicolumn{2}{c}{\textbf{MedCalc.}}
& \multicolumn{2}{c}{\textbf{BigCode.}}
& \multicolumn{2}{c}{\textbf{Overall}} \\
\cmidrule(lr){2-3}
\cmidrule(lr){4-5}
\cmidrule(lr){6-7}
\cmidrule(lr){8-9}
\cmidrule(lr){10-11}
\cmidrule(lr){12-13}
\cmidrule(lr){14-15}
& \textbf{R@1} & \textbf{R@10}
& \textbf{R@1} & \textbf{R@10}
& \textbf{R@1} & \textbf{R@10}
& \textbf{R@1} & \textbf{R@10}
& \textbf{R@1} & \textbf{R@10}
& \textbf{R@1} & \textbf{R@10}
& \textbf{R@1} & \textbf{R@10} \\
\midrule

Ours
& 74.7 & 93.8 & 13.8 & 54.0 & 41.4 & 68.2 & 17.8 & 56.1 & 93.6 & 98.1 & 23.0 & 61.1 & 47.9 & 73.8 \\

\midrule
\rowcolor{retrievalblue}
\multicolumn{15}{c}{\textbf{\textit{Indexing}}} \\
\rowcolor{retrievalblue}
w/~~Graph $\mathcal{G}$
& 58.9 & 74.6
& 15.3 & 51.3
& 8.8 & 14.1
& 15.4 & 50.4
& 70.4 & 80.5
& 10.8 & 25.8
& 29.9 & 45.3 \\
\rowcolor{retrievalblue}
w/~~Tree $\mathcal{T}$
& 24.2 & 29.2
& 4.5 & 11.7
& 4.7 & 5.6
& 7.8 & 21.6
& 36.1 & 36.9
& 4.3 & 7.9
& 13.8 & 17.3 \\
\midrule
\rowcolor{rerankpurple}
\multicolumn{15}{c}{\textit{\textbf{Retrieval}}} \\
\rowcolor{rerankpurple}
w/o $\mathcal{A}_{\mathrm{skill}}$
& 47.7 & 56.9
& 11.3 & 29.6
& 24.9 & 39.8
& 17.3 & 43.2
& 65.9 & 72.2
& 15.8 & 30.2
& 32.2 & 45.4 \\
\rowcolor{rerankpurple}
w/o $\mathcal{A}_{\mathrm{horiz}}$
& 66.8 & 84.4
& 14.3 & 52.1
& 33.8 & 59.0
& 15.5 & 44.8
& 82.5 & 87.5
& 17.3 & 50.7
& 41.3 & 65.0 \\
\rowcolor{rerankpurple}
w/o $\mathcal{A}_{\mathrm{prune}}$
& 74.1 & 89.2
& 14.3 & 50.5
& 42.5 & 70.2
& 17.5 & 51.6
& 90.0 & 93.1
& 20.5 & 58.4
& 46.9 & 71.5 \\

\rowcolor{rerankpurple}
w/o Gate $G$
& 64.8 & 85.6
& 12.1 & 36.3
& 34.5 & 58.2
& 14.4 & 48.6
& 87.8 & 95.2
& 20.8 & 60.2
& 42.6 & 66.4 \\

\rowcolor{rerankpurple}
w/o Reward
& 74.2 & 90.0
& 14.2 & 53.3
& 41.5 & 67.3
& 17.2 & 51.2
& 89.4 & 93.4
& 20.3 & 56.7
& 46.5 & 70.9 \\

\rowcolor{rerankpurple}
w/o $\mathcal{V}_{\mathrm{hier}}$
& 74.0 & 89.7
& 14.7 & 51.6
& 41.6 & 62.1
& 17.3 & 51.6
& 87.0 & 90.9
& 20.3 & 57.4
& 46.0 & 68.9 \\

\rowcolor{rerankpurple}
w/o $\mathcal{V}_{\mathrm{visit}}$
& 74.0 & 89.6
& 14.2 & 51.2
& 41.2 & 65.4
& 16.5 & 49.3
& 87.0 & 90.9
& 19.8 & 51.2
& 45.7 & 68.3 \\
\bottomrule
\end{tabularx}
\end{table*}

\subsection{Detail Analysis}
\label{sec:analysis}
Next, we conduct a detailed analysis of the key components and behaviors of our framework.

\textbf{Effectiveness of Gate-aware MCTS Actions.} Our gate-aware MCTS includes four actions: $\mathcal{A}_{\mathrm{node}}$, $\mathcal{A}_{\mathrm{skill}}$, $\mathcal{A}_{\mathrm{horiz}}$, and $\mathcal{A}_{\mathrm{prune}}$, where the latter three extend vanilla MCTS. To evaluate their effectiveness, we analyze these actions in Table \ref{tab:rq}. Level denotes the tree depth from L0 to L4; Nodes is the number of nodes at each level; Avg. Skills is the average number of skills per node; Avg. Child is the average number of children per non-leaf node; and Avg. Sim measures the average similarity among sibling nodes based on their representations $t=(emb_t,vec_t,\mathcal{L}_t)$. $\mathcal{A}_{\mathrm{node}}$ Fail denotes the fraction of queries that are first routed incorrectly at the current level. $\mathcal{A}_{\mathrm{skill}}$ Rescue and $\mathcal{A}_{\mathrm{horiz}}$ Rescue denote the fractions of these failures recovered by applying $\mathcal{A}_{\mathrm{skill}}$ from the parent node and $\mathcal{A}_{\mathrm{horiz}}$ from the failed node, respectively. $\mathcal{A}_{\mathrm{prune}}$ Frequency denotes the proportion of selected pruning actions, while $\mathcal{A}_{\mathrm{prune}}$ Acc measures the proportion that correctly remove branches without the gold skill.
From the results, we make four observations. First, the hierarchy progressively narrows the retrieval space. The average number of skills per node decreases from 625.29 at L1 to 12.01 at L4, while Avg. Child drops from 42.00 to 3.21. Meanwhile, Avg. Sim increases from 0.22 to 0.43, indicating that deeper levels contain fewer but more fine-grained and similar nodes. Second, $\mathcal{A}_{\mathrm{node}}$ failures mainly arise from the large search width at shallow levels and fine-grained ambiguity at deeper levels. The failure rate is highest at L1 (30.56\%), decreases at L2 and L3 as the search space narrows, and rises again to 20.22\% at L4, where sibling nodes become more similar. Third, $\mathcal{A}_{\mathrm{skill}}$ and $\mathcal{A}_{\mathrm{horiz}}$ effectively recover errors introduced by $\mathcal{A}_{\mathrm{node}}$. Once $\mathcal{A}_{\mathrm{node}}$ routes to an incorrect branch, $\mathcal{A}_{\mathrm{skill}}$ and $\mathcal{A}_{\mathrm{horiz}}$ successfully redirect the search to a gold-containing branch in 41.67\%-74.51\% and 35.15\%-64.63\% of the failed cases, respectively, demonstrating their complementary roles in branch refinement and cross-branch correction. Finally, $\mathcal{A}_{\mathrm{prune}}$ is conservative but reliable. It is selected in only 2.35\%-6.45\% of cases, while $\mathcal{A}_{\mathrm{prune}}$ Acc remains consistently high at 94.97\%-96.19\%.

\begin{table*}[t]
\centering
\caption{Hierachy statics and level-wise analysis of actions.}
\label{tab:rq}
\scriptsize
\renewcommand{\arraystretch}{1.08}
\setlength{\tabcolsep}{2.8pt}

\begin{tabular*}{\textwidth}{
@{\extracolsep{\fill}}
c
c c c c
@{\hspace{10pt}}
c c c
@{\hspace{10pt}}
c c
@{}
}
\toprule
\multicolumn{5}{c}{\textbf{Hierarchy}}
& \multicolumn{3}{c}{\textbf{Action Recovery}}
& \multicolumn{2}{c}{\textbf{Prune Safety}} \\
\cmidrule(lr){1-5}
\cmidrule(lr){6-8}
\cmidrule(lr){9-10}

\textbf{Level}
& \textbf{Nodes}
& \textbf{Avg. Skills}
& \textbf{Avg. Chid}
& \textbf{Avg. Sim}
& $\boldsymbol{\mathcal{A}_{\mathrm{node}}}$ \textbf{Fail}
& $\boldsymbol{\mathcal{A}_{\mathrm{skill}}}$ \textbf{Res.}
& $\boldsymbol{\mathcal{A}_{\mathrm{horiz}}}$ \textbf{Res.}
& \textbf{$\boldsymbol{\mathcal{A}_{\mathrm{prune}}}$ Freq.}
& \textbf{$\boldsymbol{\mathcal{A}_{\mathrm{prune}}}$ Acc.} \\
\midrule

L0 & 1  & 26,262.00 & 42.00 & --   & 0.00\%  & 0.00\%  & 0.00\%  & 0.00\% & -- \\
L1 &  42   & 625.29    & 8.67  & 0.22 & 30.56\% & 64.85\% & 35.15\% & 2.53\% & 95.49\% \\
L2 & 338   & 77.52     & 4.30  & 0.32 & 15.19\% & 47.56\% & 64.63\% & 2.35\% & 95.46\% \\
L3 & 1,126 & 21.17     & 3.21  & 0.41 & 10.16\% & 74.51\% & 58.82\% & 4.48\% & 94.97\% \\
L4 & 880   & 12.01     & 0.00  & 0.43 & 20.22\% & 41.67\% & 47.22\% & 6.45\% & 96.19\% \\

\bottomrule
\end{tabular*}
\end{table*}





\begin{figure*}[t]
    \centering
    \includegraphics[width=\textwidth]{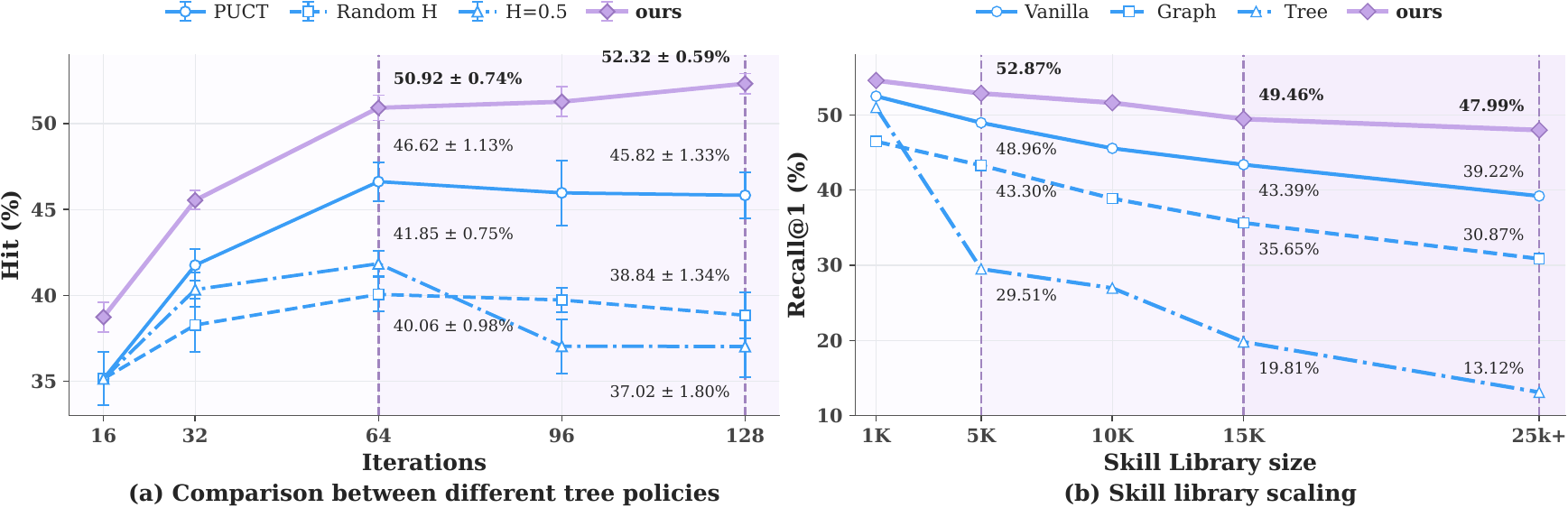}
    \caption{
    Gate effectiveness (a) and skill-library (b) scalability.}
    
    \vspace{-18pt}
    \label{figs:2_3}
\end{figure*}

\textbf{Effectiveness of the Gate in G-PUCT.}
We further examine whether the gate improves action selection and whether the node-specific entropy $H$ provides useful search signals. We record node states at iterations 16, 32, 64, 96, and 128, and compare four strategies from each identical state: PUCT without the gate, G-PUCT with the observed $Q$ and $H$, Random H with randomized entropy, and H=0.5 with fixed entropy. An action is counted as a hit if its target node contains the gold skill. All methods are evaluated on the same valid node set and repeated with five MCTS seeds, reporting mean and standard deviation. The results are reported in Figure \ref{figs:2_3}(a).
From the results, we make two observations. First, G-PUCT consistently outperforms PUCT, with the gap widening as search progresses. At iteration 64, G-PUCT achieves 50.92\% Hit versus 46.62\% for PUCT; at iteration 128, the gap grows to 6.50\% (52.32\% vs. 45.82\%). This indicates that the gate increasingly benefits from accumulated search statistics. Second, the observed entropy $H$ is critical for effective gating. At iteration 128, Random H and H=0.5 achieve only 38.84\% and 37.02\%, respectively, far below G-PUCT. Thus, the gains arise not merely from adding a gate term, but from using node-specific uncertainty to adapt action preferences. 
\textbf{Scalability to large skill libraries.}
To evaluate the scalability of different retrieval methods, we progressively enlarge the skill library from 1K to 5K, 10K, 15K, and the full 26,262 skills, and report Recall@1 at each scale. As the library grows, more irrelevant skills are introduced, making it increasingly difficult to identify the most relevant skill. We compare four methods: Vanilla, which ranks skills by weighted similarity; Graph, which performs PPR over the global skill graph; 
Tree, which applies MCTS over the hierarchical tree; and Ours, which combines hierarchical routing with graph-based search. As shown in Figure~\ref{figs:2_3}(b), all methods degrade as the skill library grows, but Ours consistently achieves the highest Recall@1 and the smallest performance drop. From 1K to the full library, Vanilla and Graph decrease by 10.38 and 15.60 percentage points, respectively, while Tree suffers a much larger drop of 37.88 points. In contrast, Ours remains relatively stable across different library sizes. These results suggest that similarity-based retrieval is increasingly affected by distractors, global graph search becomes more susceptible to noise as the graph expands, and tree-only search tends to accumulate routing errors. By progressively narrowing the search space through the hierarchy and using graph actions to recover from incorrect routes, our full framework scales more robustly to large skill libraries.

\section{Conclusion}
We introduced SkillGATE, a graph-guided hierarchical framework for skill retrieval with Gate-Aware MCTS. Inspired by Optimal Foraging Theory, SkillGATE formulates retrieval as an adaptive process that coordinates region-level navigation with skill-level selection over a graph-preserving hierarchy. Experiments demonstrate consistent improvements across retrieval and reranking backbones, while analyses validate the effectiveness of the proposed search mechanism. 

\section*{AI Use Statement}
We used generative AI tools to assist with standardizing mathematical notation and the presentation of equations, testing and evaluating code, and improving the clarity and readability of the manuscript. The modeling approach and core methodology were developed by the authors without generative AI assistance. The authors reviewed all AI-assisted material, checked the mathematical expressions, and verified the code evaluation results. We take full responsibility for the accuracy and integrity of the final paper, including all text, claims, and results.

\bibliography{iclr2025_conference}
\bibliographystyle{iclr2025_conference}

\newpage

\appendix
\raggedbottom
\definecolor{oursblue}{HTML}{2563EB}

\section{Algorithm}
\label{app:skillgate_algorithm}
In this section, we present the algorithms for \textit{offline indexing} (Algorithm~\ref{alg:skillgate_index}) and \textit{online retrieval} (Algorithm~\ref{alg:skillgate_retrieve}).

\begin{algorithm}[!htbp]
\caption{\textsc{Offline Indexing}}
\label{alg:skillgate_index}
\small
\begin{algorithmic}[1]

\Require Skill library $\mathcal{S}$;
         neighborhood size $K_{\mathrm{nbr}}$;
         edge threshold $\tau_e$;
         maximum depth $l_{\max}$;
         minimum node size $n_{\min}$
\Ensure Skill graph $\mathcal{G}$ and hierarchical index $\mathcal{T}$

\Function{BuildIndex}{$\mathcal{S}$}

    \State $\mathcal{X}\gets
           \Call{Encode-Skills}{\mathcal{S}}$
           \Comment{Labels, embeddings, and lexical vectors}

    \State $\mathcal{G}\gets
           \textcolor{oursbluetext}{\textbf{Graph Construction}}
           (\mathcal{X},K_{\mathrm{nbr}},\tau_e)$
           \Comment{Weighted skill graph}

    \State $\mathcal{T}\gets
           \textcolor{oursbluetext}{\textbf{Tree Construction}}
           (\mathcal{G},l_{\max},n_{\min})$
           \Comment{Recursive Leiden decomposition}

    \For{$u\in\mathcal{T}$}

        \State $(\mathrm{emb}_u,\mathrm{vec}_u,\mathcal{L}_u)
               \gets
               \Call{Aggregate}{\operatorname{Skills}(u)}$

    \EndFor

    \State \Return $(\mathcal{G},\mathcal{T})$

\EndFunction

\end{algorithmic}
\end{algorithm}
\begin{algorithm}[!htbp]
\caption{\textsc{Online Retrieval}}
\label{alg:skillgate_retrieve}
\small
\begin{algorithmic}[1]

\Require Query $q$;
         skill graph $\mathcal{G}$;
         hierarchical index $\mathcal{T}$;
         iteration budget $T$;
         rollout budget $B$;
         reward cutoffs $k_1,k_2$;
         output size $K$
\Ensure Retrieved skills $\mathcal{S}^{*}_K(q)$

\Function{Retrieve}{$q,\mathcal{G},\mathcal{T},T,B,K$}

    \State $\mathcal{S}^{*}(q)\gets\varnothing$;
           $\mathcal{T}_{\mathrm{search}}
           \gets\{\operatorname{Root}(\mathcal{T})\}$

    \State $N(v),M(v)\gets 0,0$;
           $Q(v)\gets 0.5$,
           $\forall v\in\mathcal{T}$
    \Comment{Neutral value before reward observation}

    \For{$t=1,\ldots,T$}

        \State $(u,a,\mathcal{V}_{\mathrm{visit}})
        \gets
        \textcolor{ourspurpletext}{\textbf{Selection}}
        (q,\mathcal{G},\mathcal{T},
        \mathcal{T}_{\mathrm{search}},N,Q)$
        \Comment{G-PUCT over expanded transitions}

        \If{$a\neq\mathcal{A}_{\mathrm{prune}}$}

            \State $v\gets
            \textcolor{ourspurpletext}{\textbf{Expansion}}
            (q,u,a,\mathcal{G},\mathcal{T})$
            \Comment{Best unexpanded successor under $a$}

            \Statex \hspace{\algorithmicindent}
            $\displaystyle
            v=
            \arg\max_{
            \substack{
            v'\in\mathcal{D}(u),\
            v'\notin\mathcal{T}_{\mathrm{search}}\\
            a\leadsto v'
            }}
            \left[
            P(v'\mid q,u)
            +
            0.4\,\kappa(u)G(u,a)
            \right]$

            \State Add $v$ and transition $(u,v)$
                   to $\mathcal{T}_{\mathrm{search}}$

            \State $u\gets v$;
                   append $u$ to $\mathcal{V}_{\mathrm{visit}}$

            \State
            \textcolor{ourspurpletext}{\textbf{Simulation}}\ $:\quad$
            $\pi\gets[\,]$
            
            \For{$\ell=1,\ldots,B$}
            
                \If{$\mathcal{D}(u)=\varnothing$}
                    \State \textbf{break}
                \EndIf
            
                \State $v\sim P(\cdot\mid q,u)$,
                       $v\in\mathcal{D}(u)$
                \Comment{Query-prior sampling}
            
                \State $a\gets\operatorname{Action}(u,v)$
            
                \State append $(u,a,v)$ to $\pi$;
                       $u\gets v$
            
            \EndFor
            
            \State $v_m\gets u$

            \State $(v_m,\pi)\gets
                   \operatorname{ReadoutFallback}(u,\pi)$
            \Comment{Ensure a valid final readout}

            \State Append rollout nodes in $\pi$
                   to $\mathcal{V}_{\mathrm{visit}}$

        \Else

            \State $v_m\gets u$
            \Comment{Terminate the current trajectory}

        \EndIf

        \State $(R(v_m),\mathcal{C}_t)
        \gets
        \Call{Evaluate}{q,v_m,r,k_1,k_2}$
        \Comment{Reward and top-$k_2$ candidates}

        \State $\mathcal{S}^{*}(q)
        \gets
        \mathcal{S}^{*}(q)\cup\mathcal{C}_t$

        \State
        \textcolor{ourspurpletext}{\textbf{Backpropagation}}\ $:\quad$
        $N(v)\gets N(v)+1$,
        $\forall v\in\mathcal{V}_{\mathrm{visit}}$
        \Comment{Actual search trajectory}

        \State $\mathcal{V}_{\mathrm{hier}}
        \gets
        \Call{AncestorPath}{v_m,\mathcal{T}}$

        \For{$v\in\mathcal{V}_{\mathrm{hier}}$}

            \State $\displaystyle
            Q(v)\gets
            \frac{M(v)Q(v)+R(v_m)}
                 {M(v)+1}$;
            $M(v)\gets M(v)+1$

        \EndFor

    \EndFor

    \State \Return
    $\operatorname{TopK}_{x\in\mathcal{S}^{*}(q)}
    r(q,x)$

\EndFunction

\end{algorithmic}
\end{algorithm}

\section{Graph-based Search Actions Details} \label{app:act}
In this section, we detail the two graph-based actions, skill-aware descend ($\mathcal{A}_{\mathrm{skill}}$) and horizontal search ($\mathcal{A}_{\mathrm{horizon}}$). Both actions share a query-guided PPR procedure but differ in their propagation scopes and target selection rules. 

\textbf{PPR Initialization.} 
Within the subgraph $\mathcal{G}_u$ associated with the current node $u$, we select the eight skill nodes with the highest query–skill relevance scores $r(q,x)$ as seeds. Let $\mathcal{S}_{\mathrm{seed}}(u)$ denote the resulting seed set. We normalize their relevance scores to obtain the restart distribution used to initialize PPR:
\begin{equation}
    p_0^u(x)=
\begin{cases}
\displaystyle
\frac{r(q,x)}
{\sum_{x'\in\mathcal{S}_{\mathrm{seed}}(u)} r(q,x')},
& x\in\mathcal{S}_{\mathrm{seed}}(u), \\[6pt]
0, & \text{otherwise}.
\end{cases}
\end{equation}

\textbf{Skill-aware Descend.}
For $\mathcal{A}_{\mathrm{skill}}$, we restrict PPR propagation to $\mathcal{G}_u$, normalizing outgoing edge weights over neighbors within this subgraph. Let $\pi_{\mathrm{intra}}(x\mid q,u)$ denote the resulting PPR score, we select the highest-scoring skill node $x^*$:
\begin{equation}
x^*=
\operatorname*{arg\,max}_{x\in\mathcal{G}_{u}(u)}
\pi_{\mathrm{intra}}(x\mid q,u),
\end{equation}
and descend to the child of $u$ whose subgraph contains $x^*$. Thus, the descent direction is determined by skill-level evidence obtained through intra-subgraph propagation.

\textbf{Horizontal Search.}
For $\mathcal{A}_{\mathrm{horizon}}$, we use the same seeds but perform PPR over the full skill graph $\mathcal{G}$, allowing relevance signals to reach external skills through cross-subgraph edges. Let $\pi_{\mathrm{inter}}(x\mid q,u)$ denote the resulting PPR score. We group retained external skills by their corresponding sibling subgraphs. For each sibling node $v\in\operatorname{Sib}(u)$, we select the target node $v^*$ by cumulative PPR score:
\begin{equation}
v^* = \operatorname*{arg\,max}_{v\in\operatorname{Sib}(u)}
\sum_{x\in\mathcal{G}_v}
\pi_{\mathrm{inter}}(x\mid q,u).
\end{equation}
To prevent oscillation between sibling nodes, we exclude nodes already visited in the current MCTS iteration from horizontal target selection.

\section{Implementation Details.} \label{app:details}
We use GPT-4o-mini as the LLM backbone with a temperature of 0.5 and TF-IDF as the hash encoder. We evaluate our framework under three embedding--reranker configurations: Ada-002 with GPT-4o-mini, SkillRet-Emb-0.6B with SkillRouter-Rerank-0.6B, and R3-Emb-0.6B with R3-Rerank-0.6B. Hyperparameter setting details are provided in Appendix \ref{app:hyper}. Following ComoRAG~\citep{wang2026comorag}, we further conduct all the analysis on a sampled set detailed in Appendix \ref{app:sampled_testset}.

\subsection{Hyperparameter settings}\label{app:hyper}
We tune the hyperparameters via grid search on a separate dataset, SkillRet~\citep{cho2026skillret}, and keep the selected configuration fixed for evaluation on SRA-Bench. Table~\ref{tab:hyperparameters} summarizes the hyperparameter settings, organized by the two main stages of \textsc{SkillGATE}: offline indexing and online retrieval. These settings cover graph and hierarchy construction, gate-aware search, reward estimation, and retrieval budgets. All other settings follow the original configurations. We further analyze the sensitivity of key hyperparameters on the sampled test subsets in the following section.

\begin{table}[H]
\centering
\scriptsize
\setlength{\tabcolsep}{4.5pt}
\renewcommand{\arraystretch}{1.08}
\caption{Hyperparameter settings.}
\label{tab:hyperparameters}
\begin{tabular}{lll}
\toprule
\textbf{Stage} & \textbf{Hyperparameter} & \textbf{Value} \\
\midrule
\multirow{4}{*}{\textbf{Index}}
& Edge weights: sem./lex./label
& 0.35/0.20/0.45 \\
& Neighbor limit $K_{\mathrm{nbr}}$ 
& 50 \\
& Edge threshold $\tau_e$
& 0.32 \\
& Depth $l_{\operatorname{max}}$ / Number $n_{\operatorname{min}}$
& 4/20 \\

\midrule
\multirow{6}{*}{\textbf{Retrieval}}
& Relevance weights: sem./lex./label
& 0.50/0.25/0.25\\
& Gate Transition scale $\tau_{\kappa}$ & 16 \\
& Entropy temperature $\tau_H$ / Prior temperature $\tau_P$
& 0.20/0.90 \\
& Reward: $k_1$/$k_2$ 
& 5/10  \\
& Reward weights: top-1/top-$k_1$/top-$k_2$
& 0.55/0.30/0.15 \\
& Maximum iterations/Final output K/Search budget
& 128/50/16\\
\bottomrule
\end{tabular}
\end{table}

\subsection{Sensitivity Analysis} 
We further investigate the sensitivity of \textsc{SkillGATE} to the query-relevance and reward weights, using Ada-002 as the embedding model without reranking. Specifically, we vary one group of weights at a time while keeping all other settings fixed. Results are reported in Table~\ref{tab:weight_sensitivity}. 

\textbf{Query-relevance weights.}
The results suggest that standalone signal strength does not directly determine its contribution to the combined relevance score. Although lexical relevance performs best among the single-signal settings (\colorbox{retrievalblue}{blue rows}), the best-performing configuration $(0.50, 0.25, 0.25)$ assigns the largest weight to semantic relevance. Moreover, all single-signal settings yield lower Recall@10 than this combination, supporting the value of integrating complementary evidence rather than relying exclusively on the strongest individual signal. The mixed-weight settings (\colorbox{rerankpurple}{purple rows}) exhibit variations of only 0.36 and 0.60 percentage points in Recall@1 and Recall@10, respectively, suggesting that this combination allows flexibility in weighting rather than requiring a narrowly tuned configuration.

\textbf{Reward weights.}
The results highlight the distinction between identifying a high-scoring skill and estimating the utility of an entire search branch. Although the top-1 term receives the largest weight in the default configuration $(0.55, 0.30, 0.15)$, it performs worst when used alone (\colorbox{retrievalblue}{blue rows}). Combining it with the top-$k_1$ and top-$k_2$ terms achieves the best performance, suggesting that the strongest individual relevance score is more useful when supported by evidence from multiple highly ranked candidates. Thus, effective reward estimation should emphasize the best candidate without treating its score as sufficient evidence of branch quality. The limited variations across mixed-weight settings (\colorbox{rerankpurple}{purple rows})---0.41 and 0.57 percentage points in Recall@1 and Recall@10, respectively---further suggest that this benefit is not highly sensitive to the precise weighting within the evaluated range.

\begingroup
\definecolor{retrievalblue}{HTML}{EFF7FE}
\definecolor{rerankpurple}{HTML}{F5F0FB}

\begin{table}[H]
\centering
\caption{Sensitivity to query-relevance and reward weights.}
\label{tab:weight_sensitivity}

\scriptsize
\setlength{\tabcolsep}{1.4pt}
\renewcommand{\arraystretch}{1.07}

\begin{tabularx}{\textwidth}{
@{}
>{\hsize=.68\hsize\centering\arraybackslash}X
>{\hsize=.68\hsize\centering\arraybackslash}X
>{\hsize=.68\hsize\centering\arraybackslash}X
>{\hsize=1.48\hsize\centering\arraybackslash}X
>{\hsize=1.48\hsize\centering\arraybackslash}X
>{\hsize=.68\hsize\centering\arraybackslash}X
>{\hsize=.68\hsize\centering\arraybackslash}X
>{\hsize=.68\hsize\centering\arraybackslash}X
>{\hsize=1.48\hsize\centering\arraybackslash}X
>{\hsize=1.48\hsize\centering\arraybackslash}X
@{}
}

\toprule

\multicolumn{5}{c}{\textbf{Query-Relevance Weights}}
&
\multicolumn{5}{c}{\textbf{Reward Weights}} \\
\cmidrule(lr){1-5}
\cmidrule(lr){6-10}

\multicolumn{3}{c}{\textbf{Relevance Weights}}
& \multicolumn{2}{c}{\textbf{Retrieval}}
&
\multicolumn{3}{c}{\textbf{Reward Weights}}
& \multicolumn{2}{c}{\textbf{Retrieval}} \\
\cmidrule(lr){1-3}
\cmidrule(lr){4-5}
\cmidrule(lr){6-8}
\cmidrule(lr){9-10}

sem. & lex. & label
& R@1 & R@10
& top-1 & top-$k_1$ & top-$k_2$
& R@1 & R@10 \\
\midrule

\textbf{0.50}
& \textbf{0.25}
& \textbf{0.25}
& $\mathbf{47.98\pm0.12}$
& $\mathbf{74.46\pm0.32}$
&
\textbf{0.55}
& \textbf{0.30}
& \textbf{0.15}
& $\mathbf{47.98\pm0.12}$
& $\mathbf{74.46\pm0.32}$ \\
\midrule

\rowcolor{retrievalblue}
0.00 & 0.00 & 1.00
& $47.47\pm0.45$ & $72.49\pm0.63$
& 0.00 & 0.00 & 1.00
& $47.51\pm0.40$ & $73.76\pm0.36$ \\

\rowcolor{retrievalblue}
0.00 & 1.00 & 0.00
& $47.54\pm0.31$ & $73.16\pm0.56$
& 0.00 & 1.00 & 0.00
& $47.22\pm0.33$ & $73.70\pm0.64$ \\

\rowcolor{retrievalblue}
1.00 & 0.00 & 0.00
& $46.83\pm0.14$ & $71.74\pm0.59$
& 1.00 & 0.00 & 0.00
& $46.80\pm0.28$ & $72.80\pm0.26$ \\
\midrule

\rowcolor{rerankpurple}
0.33 & 0.33 & 0.33
& $47.47\pm0.28$ & $73.45\pm0.56$
& 0.33 & 0.33 & 0.33
& $47.22\pm0.41$ & $73.64\pm0.52$ \\

\rowcolor{rerankpurple}
0.25 & 0.50 & 0.25
& $47.15\pm0.44$ & $73.32\pm0.41$
& 0.25 & 0.50 & 0.25
& $47.27\pm0.34$ & $73.87\pm0.18$ \\

\rowcolor{rerankpurple}
0.25 & 0.25 & 0.50
& $47.51\pm0.42$ & $73.92\pm0.49$
& 0.25 & 0.25 & 0.50
& $46.86\pm0.51$ & $73.30\pm0.36$ \\

\bottomrule
\end{tabularx}

\end{table}
\endgroup

\section{Sampled Set Details}
\label{app:sampled_testset}

\begin{wraptable}{r}{0.45\columnwidth}
\vspace{-10pt}
\centering
\scriptsize
\setlength{\tabcolsep}{3.2pt}
\renewcommand{\arraystretch}{1.05}

\caption{Sampled test set statistics.}
\label{tab:sampled_testset_details}
\vspace{-3pt}

\begin{tabular}{@{}lrrrr@{}}
\toprule
\textbf{Dataset}
& \textbf{Query}
& \textbf{Gold}
& \textbf{R@1}
& \textbf{R@10} \\
\midrule
BigCodeBench & 114 & 311 & 23.92 & 63.08 \\
CHAMP        & 22  & 36  & 12.88 & 41.67 \\
LogicBench   & 76  & 76  & 14.47 & 56.58 \\
MedCalcBench & 110 & 110 & 93.64 & 97.27 \\
TheoremQA    & 75  & 75  & 68.00 & 90.67 \\
ToolQA       & 143 & 143 & 44.76 & 72.03 \\
\midrule
\textbf{Overall}
& \textbf{540}
& \textbf{751}
& \textbf{47.98}
& \textbf{74.46} \\
\bottomrule
\end{tabular}

\vspace{-10pt}
\end{wraptable} 

We randomly sample 10\% of the queries from each of the six benchmarks using a fixed random seed, yielding a subset of 540 queries. Only the query set is subsampled; the complete library of 26,262 skills remains unchanged. To assess whether this subset yields performance comparable to the full benchmark, we report the results of \textsc{SkillGATE} on the subset using Ada-002 without reranking. As Tabel \ref{tab:sampled_testset_details} shown, \textsc{SkillGATE} achieves 47.98\% R@1 and 74.46\% R@10 on the subset, compared with 47.88\% and 73.83\%, respectively, on the full benchmark (Table~\ref{tab:main_results}). The respective differences of 0.10 \% and 0.63\% indicate comparable SR performance between the sampled subset and the full benchmark.

\section{Analysis of Search Strategies}
To evaluate the effectiveness of our search strategy, we compare G-PUCT with six alternatives on the same index. The \colorbox{retrievalblue}{blue rows} correspond to graph propagation methods: Random Walk (RW)~\citep{lovasz1993random}, which samples transitions between neighboring skill nodes according to the graph's transition probabilities; Heat Kernel Diffusion (HKD)~\citep{chung2007heat}, which aggregates neighborhood information across different propagation lengths through heat-kernel diffusion; and Personalized PageRank (PPR)~\citep{jeh2003scaling}, which incorporates restarts to the seed distribution to maintain the influence of seed nodes during propagation.
The \colorbox{rerankpurple}{purple rows} correspond to hierachy-based search strategies: Greedy~\citep{pearl1984heuristics}, which selects the child node with the highest prior probability $P(v)$ at each step; UCT~\citep{kocsis2006bandit}, which combines historical mean returns with a visit-count-based exploration bonus to balance exploiting promising branches and exploring less-visited ones; and PUCT~\citep{silver2016mastering}, which additionally weights the exploration bonus by the prior probability $P(v)$ to guide node selection.

From Table \ref{tab:search_strategy}, we yields three observations: 1) Improving graph propagation alone yields limited gains, highlighting the value of hierarchical guidance.
RW, HKD, and PPR achieve similar performance: replacing RW with PPR improves R@1/R@10 by only 0.99/1.24 percentage points. In contrast, even without search feedback, hierarchy-based Greedy achieves 36.93/59.80, outperforming PPR at 31.04/47.82. These results suggest that the limitations of graph-only retrieval cannot be fully addressed by changing the propagation rule alone; organizing the search space through hierarchical guidance is also important.
2) Combining hierarchical guidance with graph relations and search feedback improves upon prior-only descent.
Both UCT and PUCT outperform Greedy, which selects children solely according to their priors. In particular, PUCT improves R@1/R@10 from 36.93/59.80 to 43.73/67.62. This suggests that hierarchical organization provides a useful starting point, but effective retrieval further benefits from incorporating relational evidence and accumulated search feedback rather than relying exclusively on node priors.
3) How actions are selected matters beyond which search components are available.
G-PUCT, UCT, and PUCT all exploit hierarchy, graph relations, and search feedback, yet G-PUCT achieves the highest R@1/R@10 of 47.98/74.46, exceeding the strongest alternative, PUCT, by 4.25/6.84 percentage points. These gains support the value of gate-aware action selection: providing access to complementary search operations is not sufficient without an effective policy for deciding when to use them. The larger improvement in R@10 further indicates that the benefits extend beyond the top-ranked result to broader coverage of relevant skills within the top ten results.

\begin{table}[H]
\centering
\scriptsize
\setlength{\tabcolsep}{3.2pt}
\renewcommand{\arraystretch}{1.08}
\caption{
Comparison of search strategies on the sampled subset.
}
\label{tab:search_strategy}

\begin{tabular}{cccccc@{\hspace{5pt}}cc}
\toprule
\textbf{Search Strategy}
& \textbf{Hierarchy}
& \textbf{Graph}
& \textbf{Feedback}
& \textbf{R@1}
& \textbf{R@10} \\
\midrule

\rowcolor{retrievalblue}
RW
& $\times$ & $\checkmark$ & $\times$ 
& 30.05 & 46.58\ \\

\rowcolor{retrievalblue}
HKD
& $\times$ & $\checkmark$ & $\times$ 
& 30.52 & 47.06 \\

\rowcolor{retrievalblue}
PPR
& $\times$ & $\checkmark$ & $\times$ 
& 31.04 & 47.82 \\

\addlinespace[1pt]
\hdashline
\addlinespace[1pt]

\rowcolor{rerankpurple}
Greedy
& $\checkmark$ & $\times$ & $\times$
& 36.93 & 59.80 \\


\rowcolor{rerankpurple}
UCT
& $\checkmark$ & $\checkmark$ & $\checkmark$ 
& 42.16 & 65.51 \\

\rowcolor{rerankpurple}
PUCT
& $\checkmark$ & $\checkmark$ & $\checkmark$
& 43.73 & 67.62 \\

\addlinespace[1pt]
\midrule
\addlinespace[1pt]

\rowcolor{ourspurple}
\textbf{G-PUCT}
& $\checkmark$ & $\checkmark$ & $\checkmark$ 
& $\mathbf{47.98}$ & $\mathbf{74.46}$ \\

\bottomrule
\end{tabular}
\end{table}









\section{In-depth Analysis of Decoupled Backpropagation}
\begin{wraptable}{r}{0.42\columnwidth}
\vspace{-1pt}
\centering
\scriptsize
\setlength{\tabcolsep}{4.5pt}
\renewcommand{\arraystretch}{1.05}

\caption{Update-path comparison.}
\label{tab:level_paths}
\vspace{-3pt}

\begin{tabular}{@{}cccc@{}}
\toprule
\textbf{Level}
& $\boldsymbol{\mathcal{V}_{\mathrm{visit}}}$ \textbf{Hit}
& $\boldsymbol{\mathcal{V}_{\mathrm{hier}}}$ \textbf{Hit}
& \textbf{Ours Hit} \\
\midrule
L0 & 100.0 & 100.0 & 100.0 \\
L1 & 90.4  & 63.0  & 90.4  \\
L2 & 64.0  & 39.2  & 66.8  \\
L3 & 40.2  & 23.7  & 42.0  \\
L4 & 28.0  & 16.3  & 32.2  \\
\bottomrule
\end{tabular}
\vspace{-10pt}
\end{wraptable}

We compare our decoupled backpropagation with two coupled variants that update both the historical return $Q$ and visit count $N$ along a single path: either the hierarchical parent path $\mathcal{V}_{\mathrm{hier}}$ or the actual visited path $\mathcal{V}_{\mathrm{visit}}$. 

From Table~\ref{tab:level_paths}, we make two observations. First, $\mathcal{V}_{\mathrm{hier}}$ yields the lowest hit rate at every non-root level, decreasing from 63.0\% at L1 to 16.3\% at L4. Its substantial performance gap relative to the other two variants suggests that visit counts should reflect the actual search trajectory rather than hierarchical ancestry. Second, $\mathcal{V}_{\mathrm{visit}}$ is more competitive and matches our method at L1 (90.4\%), but decoupled backpropagation achieves higher hit rates at all subsequent levels. Specifically, our method improves over $\mathcal{V}_{\mathrm{visit}}$ by 2.8, 1.8, and 4.2 percentage points at L2, L3, and L4, respectively, with the largest improvement at the deepest level. This suggests that faithfully tracking visits alone is insufficient: reward attribution also benefits from following the hierarchy of the final node. Together, these results support separating visitation tracking from reward attribution rather than updating both statistics along a single path.

\section{In-depth Analysis of Efficiency}
We compare \textsc{SkillGATE} with the graph-based SR method GoS~\citep{liu2026graphofskills} and the hierarchy-based SR method AgentSkillOS~\citep{li2026organizing} in terms of retrieval performance, token usage, and runtime. Table~\ref{tab:method_efficiency} summarizes the results. Under \textbf{Average Time Taken}, we report the average indexing time per skill and retrieval time per query, both in seconds. Under \textbf{Token Usage}, we separately report the tokens consumed by LLM calls and those processed by embedding models, both in millions. We make two observations.
\begin{table}[h]
\centering
\scriptsize
\caption{Comparison of efficiency between Structured SR methods.}
\label{tab:method_efficiency}
\setlength{\tabcolsep}{10pt}
\renewcommand{\arraystretch}{1.15}

\begin{tabular}{lccc}
\toprule
& \textbf{Ours} & \textbf{GoS} & \textbf{AgentSkillOS} \\
\midrule

\multicolumn{4}{l}{\textbf{Performance Metrics}} \\[-2pt]
\hspace{1em} R@1
& 47.88 (100.00\%)
& 28.07 (58.63\%)
& 24.30 (50.75\%)\\

\hspace{1em} R@10
& 73.83 (100.00\%)
& 48.69 (65.95\%)
& 29.66 (40.17\%)\\

\midrule

\multicolumn{4}{l}{\textbf{Token Usage (/M)}} \\[-2pt]
\hspace{1em} LLM
& 56.08 (100.00\%)
& 98.14 (175.00\%)
& 848.21 (1512.50\%)\\

\hspace{1em} Embedding
& 59.58 (100.00\%)
& 38.85 (65.21\%)
& 0.00 (0.00\%)\\

\midrule

\multicolumn{4}{l}{\textbf{Average Time Taken (/s)}} \\[-2pt]
\hspace{1em} Index
& 0.15 (100.00\%)
& 0.40 (266.67\%)
& 0.81 (540.00\%)\\

\hspace{1em} Retrieval
& 3.09 (100.00\%)
& 0.63 (20.38\%)
& 104.86 (3393.53\%)\\

\bottomrule
\end{tabular}
\end{table}
From the tabel, we make two observations. 
1) \textsc{SkillGATE} achieves the highest R@1/R@10 of 47.88/73.83 among the compared methods, using 56.08M LLM tokens and 59.58M embedding tokens. Notably, its LLM token consumption is 42.86\% lower than that of GoS and 93.39\% lower than that of AgentSkillOS. These results demonstrate that improved retrieval quality need not come at the expense of increased LLM token consumption: \textsc{SkillGATE} delivers stronger retrieval performance with substantially lower LLM token requirements.
2) The joint graph-tree index combines low construction overhead with strong retrieval performance. \textsc{SkillGATE} requires only 0.15 seconds per skill for indexing, compared with 0.40 seconds for GoS and 0.81 seconds for AgentSkillOS. For online retrieval, it achieves higher recall than AgentSkillOS while reducing the average time per query from 104.86 to 3.09 seconds, corresponding to a 33.9-fold speedup. Compared with GoS, it improves R@1/R@10 by 19.81/25.14 percentage points with an additional 2.46 seconds per query. Overall, \textsc{SkillGATE} achieves the lowest indexing time among the compared methods and substantially improves retrieval quality while keeping average retrieval latency at 3.09 seconds.

\section{Case Study}
\begingroup
\definecolor{caseblue}{HTML}{E4ECFA}
\definecolor{casehit}{HTML}{F7F1FF}
\providecolor{ourspurpletext}{HTML}{73529B}
\begin{table}[H]
\centering
\small
\setlength{\tabcolsep}{5pt}
\renewcommand{\arraystretch}{1.12}
\caption{
Case study comparing the retrieval traces of Vanilla MCTS and \textsc{SkillGATE}.
}
\label{tab:retrieval_gold_case}
\begin{minipage}{\linewidth}
\begin{center}\textbf{Input Task}\end{center}
\vspace{-3pt}\hrule\vspace{5pt}
\textbf{Query:} What is the average speed of WN1507 from COS to MDW (miles per minute)?\par\vspace{2pt}
\textbf{Gold skill:} Flight Data: Multi-Record Aggregation (\texttt{toolqa\_003}).\par
\vspace{5pt}\hrule

\begin{center}\textbf{Search Trace of Vanilla MCTS}\end{center}
\vspace{-3pt}{\footnotesize\setlength{\tabcolsep}{3.8pt}
\begin{tabularx}{\linewidth}{
@{}
r
l
X
l
X
l
X
r
c
@{}
}
\toprule
\#
& \textbf{Phase}
&
& \textbf{Operation}
&
& \textbf{Scope Transition}
&
& \textbf{Size}
& \textbf{Gold?} \\
\midrule
1 & Selection & & Node-aware descent & & \texttt{root} $\rightarrow$ \texttt{p310} & & 1214 & No \\
2 & Expansion & & Skill-aware descent & & \texttt{p310} $\rightarrow$ \texttt{p387} & & 61 & No \\
\rowcolor{caseblue}
3 & Simulation & & Skill-aware descent & & \texttt{p387} $\rightarrow$ \texttt{p22} & & 48 & No \\
\rowcolor{caseblue}
4 & Simulation & & Node-aware descent & & \texttt{p22} $\rightarrow$ \texttt{p24} & & 17 & No \\
\rowcolor{caseblue}
5 & Evaluation & & Local candidate readouts & & \texttt{p24} $\rightarrow$ \texttt{p24} & & 17 & No \\
\bottomrule
\end{tabularx}
}

\vspace{5pt}
\begin{center}\textbf{Search Trace of SkillGate}\end{center}
\vspace{-3pt}{\footnotesize\setlength{\tabcolsep}{3.8pt}
\begin{tabularx}{\linewidth}{
@{}
r
l
X
l
X
l
X
r
c
@{}
}
\toprule
\#
& \textbf{Phase}
&
& \textbf{Operation}
&
& \textbf{Scope Transition}
&
& \textbf{Size}
& \textbf{Gold?} \\
\midrule
1 & Selection & & Node-aware descent & & \texttt{root} $\rightarrow$ \texttt{p310} & & 1,214 & No \\
2 & Expansion & & Skill-aware descent & & \texttt{p310} $\rightarrow$ \texttt{p387} & & 61 & No \\
\rowcolor{casehit}
3 & Simulation & & Horizontal search & & \texttt{p387} $\rightarrow$ \texttt{p1734} & & 17 & No \\
\rowcolor{casehit}
4 & Simulation & & Horizontal search & & \texttt{p1734} $\rightarrow$ \texttt{p1732} & & 21 & \textbf{Yes} \\
\rowcolor{casehit}
5 & Evaluation & & Candidate collection & & \texttt{p1732} $\rightarrow$ \texttt{p1732} & & 21 & \textbf{Yes} \\
\bottomrule
\end{tabularx}
}

\vspace{6pt}

\begin{center}
\textbf{Examples of Skills in Visited Nodes}
\end{center}

{\footnotesize
\setlength{\tabcolsep}{7pt}
\renewcommand{\arraystretch}{1.05}

\begin{tabularx}{\linewidth}{
@{}
p{0.11\linewidth}
p{0.30\linewidth}
>{\raggedright\arraybackslash}X
@{}
}
\toprule
\textbf{Node} & \textbf{Skill ID} & \textbf{Skill Title} \\
\midrule

\texttt{p310} & \texttt{web\_07745} & Geographic Distance: Coordinate-Based Calculation \\
& \texttt{bigcodebench\_000} & GeoPy: Distance Calculation and Geocoding \\
& \texttt{web\_11953} & Astropy: Astronomical Data Analysis \\
\addlinespace[1.5pt]

\texttt{p387} & \texttt{web\_07745} & Geographic Distance: Coordinate-Based Calculation \\
& \texttt{bigcodebench\_000} & GeoPy: Distance Calculation and Geocoding \\
& \texttt{web\_06171} & PostGIS: Spatial Queries and Routing \\
\addlinespace[1.5pt]

\rowcolor{caseblue}
\texttt{p22} & \texttt{theoremqa\_073} & Wave Mechanics: Speed, Frequency, and Wavelength \\
\rowcolor{caseblue}
& \texttt{theoremqa\_302} & Statistical Physics: Kinetic Theory and Fermi Statistics \\
\rowcolor{caseblue}
& \texttt{theoremqa\_016} & Physical Pendulum: Oscillation Period Calculation \\
\addlinespace[1.5pt]

\rowcolor{caseblue}
\texttt{p24} & \textcolor{oursbluetext}{\texttt{theoremqa\_073}} & \textcolor{oursbluetext}{Wave Mechanics: Speed, Frequency, and Wavelength} \\
\rowcolor{caseblue}
& \textcolor{oursbluetext}{\texttt{theoremqa\_302}} & \textcolor{oursbluetext}{Statistical Physics: Kinetic Theory and Fermi Statistics} \\
\rowcolor{caseblue}
& \textcolor{oursbluetext}{\texttt{theoremqa\_026}} & \textcolor{oursbluetext}{Shock Waves: Mach Cone and Sonic Boom Arrival Time} \\
\addlinespace[1.5pt]

\rowcolor{casehit}
\texttt{p1734} & \texttt{web\_02816} & Google Maps: Place Search and Navigation \\
\rowcolor{casehit}
& \texttt{web\_25548} & Google Maps CLI: Place Search and Route Planning \\
\rowcolor{casehit}
& \texttt{web\_10840} & Location Search: Natural-Language Place Queries \\
\addlinespace[1.5pt]
\rowcolor{casehit}
\texttt{p1732} & \texttt{toolqa\_002} & Flight Data: Single-Record Lookup \\
\rowcolor{casehit}
& \texttt{web\_10302} & Route Planning: Distance, Duration, and Directions \\
\rowcolor{casehit}
&
\textcolor{ourspurpletext}{\textbf{\texttt{toolqa\_003}}} &
\textcolor{ourspurpletext}{\textbf{Flight Data: Multi-Record Aggregation}} \\
\bottomrule
\end{tabularx}
}
\vspace{5pt}

\end{minipage}
\end{table}
\endgroup

\section{Prompt Templates}
\vspace{-2pt}

\begin{tcolorbox}[
appendixprompt,
breakable,
width=\textwidth,
colback=retrievalblue,
colframe=retrievalpromptframe,
colbacktitle=retrievalpromptframe!85!white,
title={Prompt Template for Structured Label Extraction}
]

\noindent\textbf{Role}\par\smallskip
You extract structured labels from executable agent skills for SkillGATE.

\par\smallskip
\noindent\textbf{Grounding Constraints}\par\smallskip
Extract only information explicitly supported by the provided skill.
Do not invent unsupported inputs, outputs, tools, APIs, or libraries.
Use empty arrays when the corresponding information is unavailable.

\par\smallskip
\noindent\textbf{Input / User Message (JSON)}\par\smallskip
{\footnotesize\ttfamily\raggedright
\{\par
"skill\_id": "\{skill\_id\}",\par
"name": "\{skill\_name\}",\par
"description": "\{skill\_description\}",\par
"content": "\{skill\_content\}"\par
\}\par
}

\par\smallskip
\noindent\textbf{Response Format}\par\smallskip
Return JSON only with the following fields:\par\smallskip
{\footnotesize\ttfamily\raggedright
\{\par
"inputs": [...],\par
"outputs": [...],\par
"tools\_apis\_libraries": [...]\par
\}\par
}

\end{tcolorbox}

\vspace{-2pt}

\begin{tcolorbox}[
appendixprompt,
breakable,
width=\textwidth,
colback=rerankpurple,
colframe=rerankpromptframe,
colbacktitle=rerankpromptframe!85!white,
title={Prompt Template for LLM-based Skill Reranking}
]

\noindent\textbf{Task}\par\smallskip
Given the following problem, rank the skills below by relevance.

\par\smallskip
\noindent\textbf{Input / User Message}\par\smallskip
{\footnotesize\ttfamily\raggedright
Problem:\par
\{query\}\par
\smallskip
Skills:\par
[1] \{skill\_name\_1\}: \{skill\_description\_1\}\par
[2] \{skill\_name\_2\}: \{skill\_description\_2\}\par
...\par
[K] \{skill\_name\_K\}: \{skill\_description\_K\}\par
}

\par\smallskip
\noindent\textbf{Response Format}\par\smallskip
Output ONLY the skill numbers in order from most to least relevant,
separated by commas.\par\smallskip
{\footnotesize\ttfamily\raggedright
Most relevant first (numbers only):\par
}

\end{tcolorbox}

\end{document}

%% file: math_commands.tex
\usepackage{amsmath,amsfonts,bm}

\def\eqref#1{equation~\ref{#1}}

\def\1{\bm{1}}

\DeclareMathAlphabet{\mathsfit}{\encodingdefault}{\sfdefault}{m}{sl}
\SetMathAlphabet{\mathsfit}{bold}{\encodingdefault}{\sfdefault}{bx}{n}

